\documentclass[pra,twocolumn,floatfix,superscriptaddress]{revtex4}

\usepackage[utf8]{inputenc} 
\usepackage[T1]{fontenc}    
\usepackage{hyperref}       
\usepackage{url}            
\usepackage{booktabs}       
\usepackage{amsfonts}       
\usepackage{nicefrac}       
\usepackage{microtype}      
\usepackage{amsmath} 
\usepackage{mathrsfs}
\usepackage[pdftex]{graphicx}
\usepackage{tikz}
\usepackage{placeins}
\usepackage{bbm}

\usepackage{mathabx}

\DeclareMathOperator{\argmax}{argmax}
\DeclareMathOperator{\tr}{tr}
\newcommand{\eff}{\mathrm{eff}}

\begin{document}

\title{Free energy-based reinforcement learning using a quantum processor}

\author{Anna~Levit}
\affiliation{1QBit,~458-550~Burrard~Street,~Vancouver~(BC),~Canada~V6C~2B5}
\author{Daniel~Crawford}
\affiliation{1QBit,~458-550~Burrard~Street,~Vancouver~(BC),~Canada~V6C~2B5}
\author{Navid~Ghadermarzy}
\affiliation{1QBit,~458-550~Burrard~Street,~Vancouver~(BC),~Canada~V6C~2B5}
\affiliation{Department of Mathematics, The University of British Columbia, 121-1984~Mathematics~Road,~Vancouver~(BC),~Canada~V6T~1Z2}
\author{Jaspreet~S.~Oberoi}
\affiliation{1QBit,~458-550~Burrard~Street,~Vancouver~(BC),~Canada~V6C~2B5}
\affiliation{School~of~Engineering~Science,~Simon~Fraser~University,~8888~University~Drive,~Burnaby~(BC),~Canada~V5A~1S6}
\author{Ehsan~Zahedinejad}
\affiliation{1QBit,~458-550~Burrard~Street,~Vancouver~(BC),~Canada~V6C~2B5}
\author{Pooya~Ronagh}
\email[Corresponding author: ]{pooya.ronagh@1qbit.com}
\affiliation{1QBit,~458-550~Burrard~Street,~Vancouver~(BC),~Canada~V6C~2B5}
\affiliation{Department of Mathematics, The University of British Columbia, 121-1984~Mathematics~Road,~Vancouver~(BC),~Canada~V6T~1Z2}

\begin{abstract}
Recent theoretical and experimental results suggest the possibility of using current and near-future quantum hardware in challenging sampling tasks. In this paper, we introduce free energy-based reinforcement learning (FERL) as an        application of quantum hardware. We propose a method for processing a quantum annealer's measured qubit spin configurations in approximating the free energy of a quantum Boltzmann machine (QBM). We then apply this method to perform reinforcement learning on the grid-world problem  
using the D-Wave 2000Q quantum annealer. The experimental results show that our technique is a promising method for harnessing the power of quantum sampling in reinforcement learning tasks.   
\end{abstract}

\maketitle

\section{Introduction} 

Reinforcement learning \cite{sutton-book, bertsekas1996neuro} has been  successfully applied  in fields such as engineering \cite{derhami2013applying, syafiie2007model}, sociology \cite{erev1998predicting, shteingart2014reinforcement}, and economics \cite{matsui2011compound, sui2010reinforcement}. 
The training samples in reinforcement learning are provided by the interaction of an agent with an ambient environment. For example, in a motion planning problem in uncharted territory, it is desirable for the agent to learn in the fastest way possible to correctly navigate using the fewest blind decisions. That is, neither exploration nor exploitation can be pursued exclusively without either facing a penalty or failing at the task. Our goal is, therefore, to not only design an algorithm that eventually converges to an optimal policy, but for the algorithm to be able to generate suboptimal policies early in the learning process. 

Free energy-based reinforcement learning (FERL) using a restricted Boltzmann machine (RBM), as suggested by \citet{hintonRBM}, relies on approximating a utility function for the agent, called the \emph{Q}-function, using the free energy of an RBM. RBMs have the advantage that their free energy can be efficiently calculated using closed formulae. RBMs can represent any joint distribution over binary variables \cite{martens2013representational, hornik1989multilayer, le2008representational}; however, this property of universality may require exponentially large RBMs \cite{martens2013representational, le2008representational}. 

\citet{2016arXiv161205695C} generalize this method by proposing the use of a quantum or quantum-inspired algorithm for efficiently approximating the free energy of a general Boltzmann machine (GBM) (in general, using GBMs involves the NP-hard problem of sampling from a Boltzmann distribution).
Using numerical simulations, they show that FERL using a deep Boltzmann machine (DBM) can provide a drastic improvement in the early stages of learning. A quantum annealer consisting of a network of quantum bits (qubits) can provide samples that approximate a Boltzmann distribution of a system of pairwise interacting qubits called the transverse-field Ising model (TFIM). The free energy of the Hamiltonian of a TFIM contains not only information about the qubits in the measurement basis, but also about their spin in a transverse direction. Using numerical simulation, \cite{2016arXiv161205695C} show that this richer many-body system can provide the same (in fact, slightly better) learning efficiency. 

In this paper, we report the results of using the D-Wave 2000Q quantum processor to experimentally verify the conjecture of \cite{2016arXiv161205695C}: the applicability of a quantum annealer in reinforcement learning. 

\section{Preliminaries} 

\subsection{Markov decision problems}\label{sec:MDP}

We refer the reader to \cite{sutton-book} and \cite{Yuksel} for an exposition on Markov decision processes (MDP), controlled Markov chains, and the various broad aspects of reinforcement learning. A \emph{Q-function} is defined by mapping a tuple $(\pi, s, a)$ of a given a \emph{stationary policy} $\pi$, a current state $s$, and an immediate action $a$ of a controlled Markov chain to the expected value of the instantaneous and future discounted rewards of the Markov chain that begins with taking action $a$ at initial state $s$ and continuing according to $\pi$: 
\begin{align*} 
Q (\pi, s, a) &= \mathbb E[ r\, (s,\, a)] + \mathbb{E} \left[ \sum\limits_{i=1}^{\infty} \gamma^{i}\, r\, (\Pi^s_{i},\, \pi(\Pi^s_{i})) \right].
\end{align*}
Here, $r(s, a)$ is a random variable, perceived by the agent from the environment, representing the immediate reward of taking action $a$ from state $s$ and $\Pi$ is the Markov chain resulting from restricting the controlled Markov chain to the policy $\pi$. The fixed real number $\gamma \in (0, 1)$ is the \emph{discount factor} of the MDP. From $Q^\ast(s, a)= \max_\pi Q(\pi, s, a)$, the optimal policy for the MDP can be retrieved via the following:
\begin{align}\label{eq:policy-from-Q}
\pi^\ast (s) = \argmax_a Q^\ast (s, a).
\end{align}
This reduces the MDP to computing $Q^\ast (s, a)$. 
Through a Bellman recursion \cite{bellman1956dynamic}, we get  
\begin{align}\label{contraction-mapping}
Q(\pi, s, a) = \mathbb E[r\, (s, a) ] + \gamma \sum_{s'} \mathbb P(s' | s, a) \max_{a'} Q (\pi, s', a'),
\end{align}
so $Q^\ast$ is the fixed point of the following operator defined on $L_\infty (S \times A)$:
$$T(Q): (s, a) \mapsto \mathbb E[ r\, (s, a)] + \gamma \int \max_{a'} Q\,.$$
In this paper, we focus on the TD(0) \emph{Q}-learning method with the \emph{Q}-function parametrized by neural networks in order to find $\pi^\ast (s)$ and $Q^\ast (s, a)$ which is based on minimizing the distance between $T(Q)$ and $Q$.

\subsection{Clamped Boltzmann machines}\label{sec:boltzmann}

A clamped Boltzmann machine is a GBM in which all visible nodes $\mathbf v$ are prescribed fixed assignments and removed from the underlying graph. Therefore, the energy of the clamped Boltzmann machine may be written as  
\begin{align}
\mathcal H_{\bold v}(\bold h)= -\sum_{v \in V,\, h \in H} w^{vh}v h -\sum_{\{h, h'\} \subseteq H} w^{hh'} hh'\,,   \label{GBMh}
\end{align}
where $V$ and $H$ are the sets of visible and hidden nodes, respectively, and by a slight abuse of notation, a letter $v$ stands both for a graph node $v \in V$ and for the assignment $v \in \{0, 1\}$. The interactions between the variables represented by their respective nodes are specified by real-valued weighted edges of the underlying undirected graph represented by $w^{vh}$, and $w^{hh'}$ denoting the weights between visible and hidden, or hidden and hidden, nodes of the Boltzmann machine, respectively. 

A clamped quantum Boltzmann machine (QBM) has the same underlying graph as a clamped GBM, but instead of a binary random variable, qubits are associated to each node of the network. The energy function is substituted by the quantum Hamiltonian of an induced TFIM, which is mathematically a Hermitian matrix
\begin{align}\label{qbm-hamiltonian}
\mathcal{H}_{\bold v}= -\sum_{v \in V,\, h \in H} w^{vh}v \sigma^z_h -\sum_{\{h, h'\} \subseteq H} w^{hh'} \sigma^z_h\sigma^z_{h'} - \Gamma \sum_{h \in H} \sigma_h^x\,,  
\end{align}
where $\sigma_h^z$ represent the Pauli $z$-matrices and $\sigma^x_h$ represent the Pauli $x$-matrices. 
Thus, a clamped QBM with $\Gamma = 0$ is equivalent to a clamped classical Boltzmann machine. This is because, in this case, $\mathcal H_{\bold v}$ is a diagonal matrix in the $\sigma^z$-basis, the spectrum of which is identical to the range of the classical Hamiltonian \eqref{GBMh}. We note that \eqref{qbm-hamiltonian} is a particular instance of a TFIM: 
\begin{align}\label{tfim}
\mathcal{H}= -\sum_{i, j} J_{i, j} \sigma_i^z \sigma_j^z - \sum_{i} h_i \sigma_i^z - \Gamma \sum_i \sigma_i^x\,.
\end{align}
The remainder of this section is formulated for the clamped QBMs, acknowledging that it can easily be specialized for clamped classical Boltzmann machines.  

\subsection{Free energy-based reinforcement learning}
\label{RL}

Let $\beta= \frac{1}{k_B T}$ be a fixed thermodynamic beta. For an assignment of visible variables $\bold v$, $F(\bold{v})$ denotes the \emph{equilibrium free energy}, and is given via 
\begin{align}\label{eq:free-def}
F(\bold{v}) := - \frac{1}\beta \ln Z_{\bold v} = \langle \mathcal H_{\bold v} \rangle + \frac{1}\beta \tr (\rho_{\bold v} \ln \rho_{\bold v})\,.
\end{align}
Here, $Z_{\bold v} = \tr (e^{-\beta \mathcal H_{\bold v}})$ is the partition function of the clamped QBM and $\rho_{\bold v}$ is the density matrix $\rho_{\bold v} = \frac{1}{Z_{\bold v}}e^{-\beta \mathcal H_{\bold v}}$. The term $- \tr (\rho_{\bold v} \ln \rho_{\bold v})$ is the entropy of the system. 
The notation $\langle \cdots \rangle$ is used for the expected value of any observable with respect to the Gibbs measure (i.e., the Boltzmann distribution), in particular,
$$\langle \mathcal H_{\bold v} \rangle = \frac{1}{Z_{\mathbf v}} \tr( \mathcal H_{\bold v} e^{-\beta \mathcal H_{\bold v}}).$$ 

Inspired by the ideas of \cite{hintonRBM} and \cite{1601.02036}, we use the negative free energy of a QBM to approximate the \mbox{\emph{Q}-function} through the relationship
$$Q (s, a) \approx -F (\mathbf s, \mathbf  a) = - F (\mathbf  s, \mathbf  a; \boldsymbol w)$$
for each admissible state--action pair \mbox{$(s, a) \in S\times A$}. Here, $\mathbf s$ and $\mathbf a$ are binary vectors encoding the state $s$ and action $a$ on the state nodes and action nodes, respectively, of a QBM. In reinforcement learning, the visible nodes of a GBM are partitioned into two subsets of state nodes $S$ and action nodes $A$. Here, $\boldsymbol w$ represents the vector of weights of a QBM as in \eqref{qbm-hamiltonian}. Each entry $w$ of $\boldsymbol w$ can now be trained using the TD(0) update rule:
\begin{align*}
\Delta w 
&= -\varepsilon (r_n({s}_n, {a}_n)
-\gamma F ({s}_{n+1}, {a}_{n+1}) + F({s}_n,{a}_n) ) \frac{\partial F}{\partial w}\,.\end{align*}
As shown in \cite{2016arXiv161205695C}, from \eqref{eq:free-def} we obtain
\begin{align}
\Delta w^{vh} &=\, \label{q-vh-update}
\varepsilon  (r_n({s}_n, {a}_n)\\  
&-\gamma F ({s}_{n+1}, {a}_{n+1}) + F({s}_n,{a}_n) ) v \langle \sigma_h^z \rangle\, \quad \text{and} \nonumber\\
\Delta w^{hh'} &=\, \label{q-hh-update}
\varepsilon (r_n({s}_n, {a}_n)\\ 
&-\gamma F ({s}_{n+1}, {a}_{n+1}) + F({s}_n,{a}_n) ) \langle \sigma_h^z\sigma_{h'}^z \rangle. \nonumber
\end{align}
This concludes the development of the REFL method using QBMs. We refer the reader to Algorithm~3 in \cite{2016arXiv161205695C} for more details. What remains to be done is to approximate values of the free energy $F(s, a)$ and also the expected values of the observables $\langle\sigma_h^z\rangle$ and $\langle \sigma_h^z \sigma_{h'}^z \rangle$. In this paper, we demonstrate how quantum annealing can be used to address this challenge.

\subsection{Adiabatic evolution of open quantum systems} 
\label{sec:open}

The evolution of a quantum system under a slowly changing time-dependent Hamiltonian is characterized by \cite{1928ZPhy}. The \emph{quantum adiabatic theorem} (QAT) in \cite{1928ZPhy} states that the system remains in its instantaneous steady state, provided there is a gap between the eigen-energy of the steady state and the rest of the Hamiltonian's spectrum at every point in time. QAT motivated \cite{2000quant.ph1106F} to introduce a paradigm of quantum computing known as quantum adiabatic computation which is closely related to the quantum analogue of simulated annealing, namely \emph{quantum annealing} (QA), introduced by \cite{PhysRevE.58.5355}. 

The history of QA and QAT inspired efforts in manufacturing physical realizations of adiabatic evolution via quantum hardware \cite{dwnature}. In reality, the manufactured chips are operated at non-zero temperature and are not isolated from their environment. Therefore, the existing adiabatic theory does not cover the behaviour of these machines. A contemporary investigation in quantum adiabatic theory was therefore initiated to study adiabaticity in open quantum systems \cite{PhysRevA.71.012331, PhysRevA.93.032118, 136726301412123016, Avron2012, 2016arXiv161201505B}. These references prove adiabatic theorems for open quantum systems under various assumptions, in particular when the quantum system is coupled to a thermal bath satisfying the Kubo--Martin--Schwinger condition, implying that the instantaneous steady state is the instantaneous Gibbs state. This work in progress shows promising opportunities to use quantum annealers to sample from the Gibbs state of a TFIM. 

In practice, due to additional complications (e.g., level crossings and gap closure, described in the references above), the samples gathered from the quantum annealer are far from the Gibbs state of the final Hamiltonian. In fact, \cite{amin2015searching} suggests that the distribution of the samples would instead correspond to an instantaneous Hamiltonian at an intermediate point in time, called the \emph{freeze-out point}. Unfortunately, this point and, consequently, the strength $\Gamma$ of the transverse field at this point, is not a priori known, and also depends on the TFIM under evolution. Our goal is to simply associate a single (average) \emph{virual} $\Gamma$ to all TFIMs constructed through the FERL. Another unknown parameter is the inverse temperature $\beta$, at which the Gibbs state, the partition function, and the free energy are attained. In a similar fashion, we wish to associate a single \emph{virtual} $\beta$ to all TFIMs encountered. 
 
\begin{figure}
  \centering
    \includegraphics[scale=0.65]{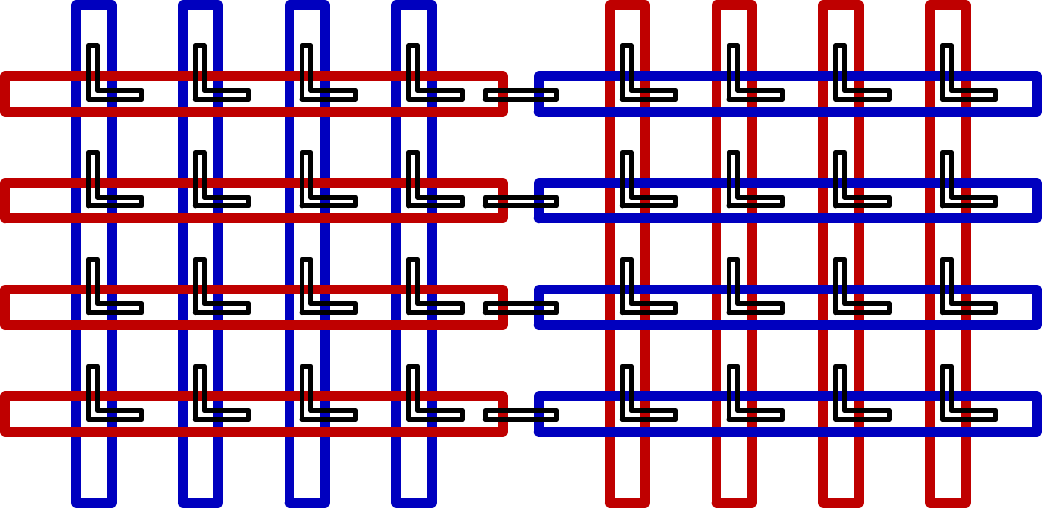} \\ \vspace{0.5cm}
    \includegraphics[width=.6\linewidth]{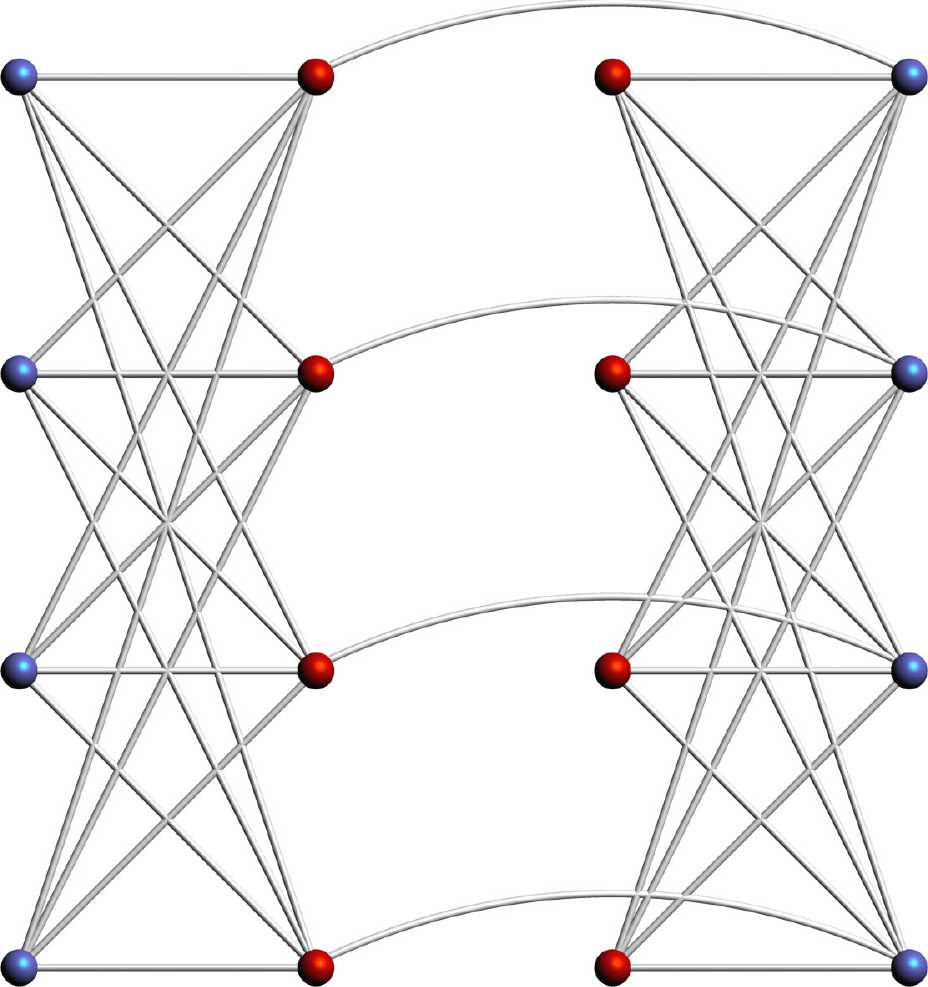}
  \caption{\footnotesize{(top) Two adjacent unit cells of the D-Wave 2000Q chip. The intra-cell couplings provide a fully connected bipartite subgraph. However, there are only four inter-cell couplings. (bottom) The Chimera graph representing the connectivity of the two unit cells of qubits.}}
  \label{fig:chimera}
\end{figure}

The quantum annealer used in our experiments is the D-Wave 2000Q, which consists of a chip of superconducting qubits connected to each other according to a sparse adjacency graph called the \emph{Chimera graph}. The Chimera graph structure looks significantly different from the frequently used models in machine learning, e.g., RBMs and DBMs, which consist of consecutive fully connected bipartite graphs. Fig \ref{fig:chimera} shows two adjacent blocks of the Chimera graph which consist of 16 qubits, which, in this paper, serve as the clamped QBM used in FERL.  

Another complication when using a quantum annealer as a QBM is that the spin configurations of the qubits can only be measured along a fixed axis (here the $z$-basis of the Bloch sphere). Once $\sigma^z$ is measured, all of the quantum information related to projection of the spin along the transverse field (i.e., the spin $\sigma^x$) collapses and cannot be retrieved. Therefore, even with a choice of virtual $\Gamma$, virtual $\beta$, and all of the measured configurations, the energy of \eqref{tfim} is still unknown. We propose a method  for overcoming this challenge based on the Suzuki--Trotter expansion of the TFIM, which we call \emph{replica stacking}, the details of which are explained in \S\ref{sec:stacking}. In \S\ref{sec:exp}, we perform a grid search over values of the virtual parameters $\beta$ and $\Gamma$. The accepted virtual parameters are the ones that result in the most effective learning for FERL in the early stages of training. 

\section{Free energy of quantum Boltzmann machines} 

\begin{figure}
  \centering
    \includegraphics[scale=0.75]{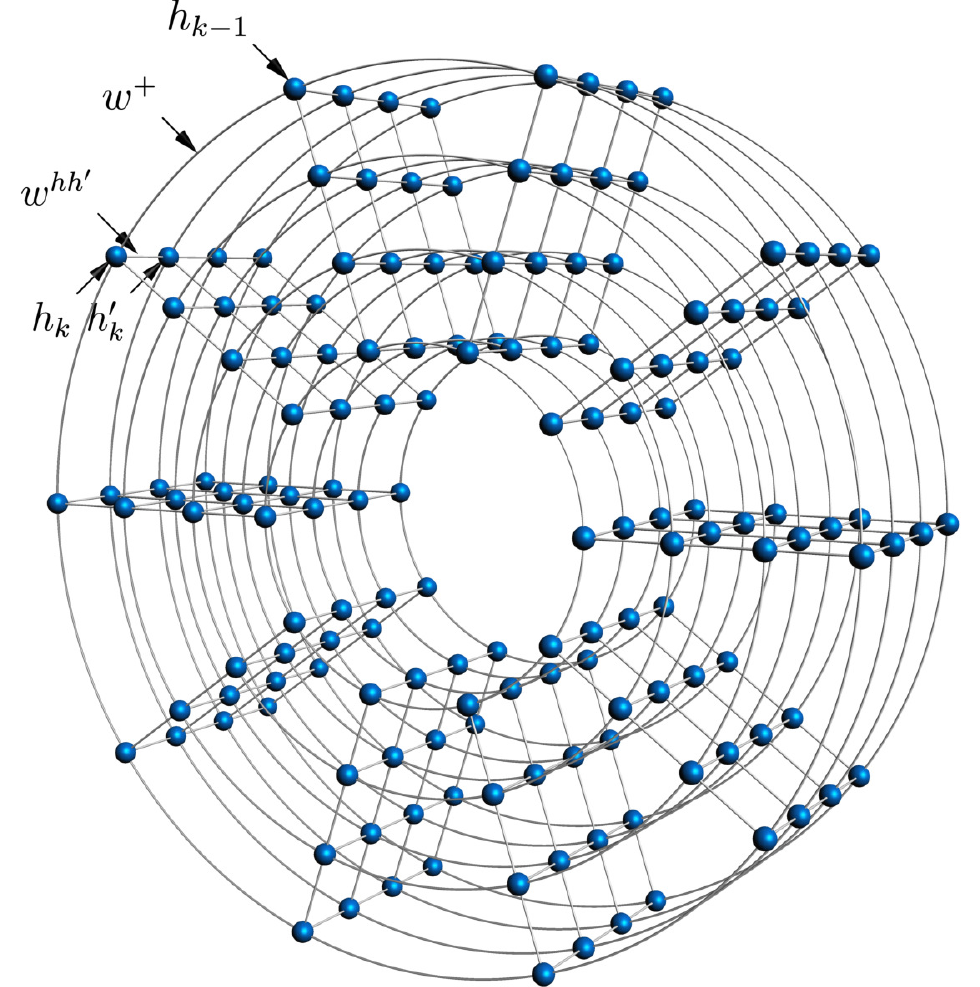} \\ \vspace{0.5cm}
    \includegraphics[scale=0.75]{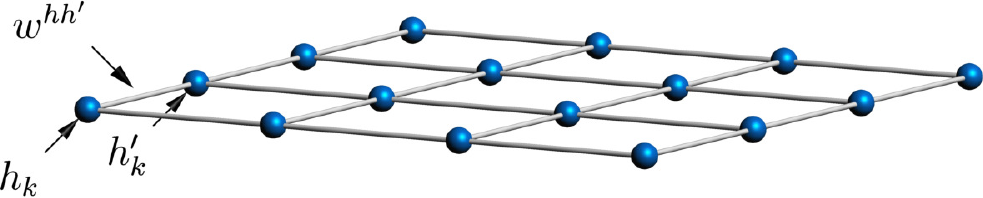}
  \caption{\footnotesize{(bottom) A transverse-field Ising model consisting of 16 qubits arranged on a two-dimensional lattice with nearest neighbour couplings. (top) The corresponding effective classical Ising model with ten replicas arranged in a three-dimensional solid torus.}}
  \label{fig:effective}
\end{figure}

\subsection{Suzuki--Trotter representation}

By the Suzuki--Trotter decomposition \cite {suzuki1976relationship}, the partition function of the TFIM defined by the Hamiltonian \eqref{qbm-hamiltonian} can be approximated using the partition function of a classical Hamiltonian denoted by $\mathcal H^\eff_{\bold v}$, which corresponds to a classical Ising model of one dimension higher. More precisely, 
\begin{align}
\label{eff-hamiltonian}
\mathcal H_{\bold v}^\eff(\bold h)&=
-\sum_{h, h'}\sum_{k=1}^{r}\frac{w^{hh'}}{r}h_{k}h'_{k}
-\sum_{v, h} \sum_{k=1}^{r}\frac{w^{vh}v }{r}h_{k}\\
&\quad -w^+\left( \sum_{h} \sum_{k=1}^{r}h_{k}h_{k+1}+\sum_{h} h_{1}h_{r}\right),\nonumber
\end{align}
where $r$ is the number of replicas, $w^+ = \frac{1}{2\beta} \log \coth \left(\frac{\Gamma\beta}{r}\right)$, and $h_{k}$ represent spins of the classical system of one dimension higher. Note that each hidden node's Pauli $z$-matrices $\sigma_h^z$ are represented by $r$ classical spins, denoted by $h_k$, with a slight abuse of notation. In other words, the original Ising model with non-zero transverse field represented through non-commuting operators can be mapped to a classical Ising model of one higher dimension. Fig. \ref{fig:effective} shows the underlying graph of a TFIM on a two-dimensional lattice and a corresponding ten-replica effective Hamiltonian in three dimensions. 

\subsection{Approximation of free energy using Gibbs sampling}

To approximate the right-hand side of each of \eqref{q-vh-update} and \eqref{q-hh-update}, we sample from the Boltzmann distribution of the effective Hamiltonian using \cite[Theorem 6]{suzuki1976relationship}. We find the expected values of the observables $\langle\sigma_h^z\rangle$ and $\langle \sigma_h^z \sigma_{h'}^z \rangle$ by averaging the corresponding classical spin values. To approximate the \emph{Q}-function, we use \cite[Theorem 4]{suzuki1976relationship} to substitute \eqref{eq:free-def} by
\begin{align}
\label{freeen-eff}
&F (\bold{v}) = \langle \mathcal H^\eff_{\bold v} \rangle + 
\frac{1}\beta
\sum_{c}\mathbb{P}(c|\bold{v}) \log \mathbb{P}(c|\bold{v})\,,
\end{align}
where $\mathcal H^\eff_{\bold v}$ is the effective Hamiltonian and $c$ ranges over all spin configurations of the classical Ising model of one dimension higher, defined by $\mathcal H^\eff_{\bold v}$.

The above argument also holds in the absence of the transverse field, that is, for the classical Boltzmann machine. In this case, the TD(0) update rule is given by

\begin{align}
\Delta w^{vh} &= \label{vh-update} 
\varepsilon  (r_n({s}_n, {a}_n)\\ 
&+\gamma Q ({s}_{n+1}, {a}_{n+1}) - Q({s}_n,{a}_n) ) v \langle h \rangle\quad \text{ and } \nonumber\\
\Delta w^{hh'} &=\label{hh-update}
\varepsilon (r_n({s}_n, {a}_n) \\
&+\gamma Q ({s}_{n+1}, {a}_{n+1}) - Q({s}_n,{a}_n) ) \langle hh' \rangle\,, \nonumber
\end{align}
where $\langle h \rangle$ and $\langle h h' \rangle$ are the expected values of the variables and the products of the variables, respectively, in the binary encoding of the hidden nodes with respect to the Boltzmann distribution of the classical Hamiltonian (\ref{GBMh}). 
%
%
The values of the \emph{Q}-functions in \eqref{vh-update} and \eqref{hh-update} can also be approximated empirically, since, in a classical Boltzmann machine, 
\begin{align}
\label{freeen}
&F (\bold{v}) = \sum_{\bold{h}}\mathbb{P}(\bold{h}|\bold{v}) \mathscr{E}_{\bold v} (\bold{h}) + \frac{1}\beta
\sum_{\bold{h}}\mathbb{P}(\bold{h}|\bold{v}) \log \mathbb{P}(\bold{h}|\bold{v})\\
&\quad= - \sum_{\substack{s \in S\\h \in H}}w^{sh}s \langle h\rangle -  \sum_{\substack{a \in A\\h \in H}} w^{ah}a \langle h\rangle - \sum_{\{h, h'\} \subseteq H} u^{hh'}\langle hh'\rangle \nonumber \\
&\quad\quad + \frac{1}\beta \sum_{\bold{h}}\mathbb{P}(\bold{h}|\bold{s}, \bold{a}) \log \mathbb{P}(\bold{h}|\bold{s}, \bold{a}). \nonumber
\end{align}

\subsection{Simulated quantum annealing}
\label{sec:sqa}

One way to sample spin values from the Boltzmann distribution of the effective Hamiltonian is to use the simulated quantum annealing algorithm (SQA) (see \cite[p. 422]{naturalcomputingalgorithms} for an introduction). SQA is one of the many flavours of quantum Monte Carlo methods, and is based on the Suzuki--Trotter expansion described above. This algorithm simulates the quantum annealing phenomena of a TFIM by slowly reducing the strength of the transverse field at finite temperature to the desired target value. 
In our implementation, we have used a single spin-flip variant of SQA with linear transverse-field schedule as in \cite{martovnak2002quantum} and \cite{1411.5693}. Experimental studies have shown similarities in the behaviour of SQA and that of quantum annealing \cite{1510.08057, 2014arXiv1409.3827A} and its physical realization \cite{arXiv:1509.02562v2, arXiv:1401.7087v2} by D-Wave Systems. 

The classical counterpart of SQA is conventional simulated annealing (SA), which is based on thermal annealing. This algorithm can be used to sample from Boltzmann distributions that correspond to an Ising spin model in the absence of a transverse field (i.e., $\Gamma = 0$ in (\ref{tfim})). Unlike with SA, it is possible to use SQA not only to approximate the Boltzmann distribution of a classical Boltzmann machine, but also that of a quantum Hamiltonian in the presence of a transverse field. This can be done by reducing the strength of transverse field to the desired value defined by the model, rather than to zero.  It has been proven by  \cite{morita2006convergence} that the spin system defined by SQA converges to the Boltzmann distribution of the effective classical Hamiltonian of one dimension higher that corresponds to the quantum Hamiltonian. Therefore, it is straightforward to use SQA to approximate the free energy in \eqref{freeen-eff} as well as the observables $\langle\sigma_h^z\rangle$ and $\langle \sigma_h^z \sigma_{h'}^z \rangle$.       

\subsection{Replica stacking}\label{sec:stacking}

As explained in \S\ref{sec:open}, a quantum annealer provides measurements of $\sigma^z$ spins for each qubit in the TFIM. The observables $\langle\sigma_h^z\rangle$ and $\langle \sigma_h^z \sigma_{h'}^z \rangle$ can therefore be approximated by averaging over the spin configurations measured by the quantum annealer. Moreover, by \cite[Theorem 6]{suzuki1976relationship} and translation invariance, each replica of the effective classical model is an approximation of the spin measurements of the TFIM in the measurement bases $\sigma^z$. 
Therefore,  a quantum annealer that operates at a given \emph{virtual} inverse temperature $\beta$ and anneals up to a \emph{virtual} transverse-field strength $\Gamma$, a $\sigma^z$-configuration sampled by the annealer may be viewed as an instance of a classical spin configuration from a replica of the classical effective Hamiltonian of one dimension higher. 

This suggests the following method to approximate the free energy from \eqref{freeen-eff} for a TFIM. We gather a pool of configurations sampled by the quantum annealer for the TFIM considered, allowing repetitions. In $r$ iterations, a spin configuration is sampled from this pool and inserted as the next replica of the effective classical Hamiltonian consisting of $r$ replicas. 
This procedure creates a pool of effective classical spin configurations $c$, which are then employed in equation \eqref{freeen-eff} in order to approximate the free energy of the TFIM empirically. 


\section{The experiments}  \label{sec:exp}

\begin{figure}
\centering
        \begin{tikzpicture}[scale=0.75]
        \draw[ultra thick] (0,0) rectangle (5,-3);
        \foreach \row in {0,...,5}{
            \foreach \column in {0,...,3} {
                \draw[dotted] (0,0) rectangle +(1*\row, -1*\column);}} 
        \node[align=left] at (0.5,-0.5) {$R$};
        \node[align=left] at (2.5,-1.5) {$W$}; 
        \node[align=left] at (2.5, -2.5) {$P$};
        \end{tikzpicture} \\ \vspace{0.1in}
        \begin{tikzpicture}[scale=0.75]
        \draw[ultra thick] (0,-3.5) rectangle (5,-6.5);
        \foreach \row in {0,...,5}{
            \foreach \column in {0,...,3} {
                \draw[dotted] (0,-3.5) rectangle +(1*\row, -1*\column);}} 
        \node[align=left] at (0.5,-4) {$\circlearrowleft$};
        \node[align=left] at (2.5,-5) {$W$};
        \node[align=left] at (0.5, -6) {$\uparrow$};
        \node[align=left] at (4.5, -4) {$\leftarrow$};
        \node[align=left] at (2.5, -6) {$\leftarrow$};
        \node[align=left] at (0.5,-5) {$\uparrow$};
        \node[align=left] at (1.5,-4) {$\leftarrow$};
        \node[align=left] at (1.4,-5) {$\leftarrow$};
        \node[align=left] at (1.5,-4.9) {$\uparrow$};
        \node[align=left] at (1.4,-6) {$\leftarrow$};
        \node[align=left] at (1.5,-5.9) {$\uparrow$};
        \node[align=left] at (2.5,-4) {$\leftarrow$};
        \node[align=left] at (3.5,-4) {$\leftarrow$};
        \node[align=left] at (3.5,-5) {$\uparrow$};
        \node[align=left] at (3.5,-6) {$\uparrow$};
        \node[align=left] at (4.4,-5) {$\leftarrow$};
        \node[align=left] at (4.5,-4.9) {$\uparrow$};
        \node[align=left] at (4.4,-6) {$\leftarrow$};
        \node[align=left] at (4.5,-5.9) {$\uparrow$};
        \end{tikzpicture}
  \label{fig:grid-det}
  \caption{\footnotesize{(top) A $3\times 5$ grid-world problem \mbox{instance} with one reward, one wall, and one penalty. (bottom) An optimal policy for this \mbox{problem} instance is a selection of directional \mbox{arrows} indicating movement directions.}}
\end{figure}
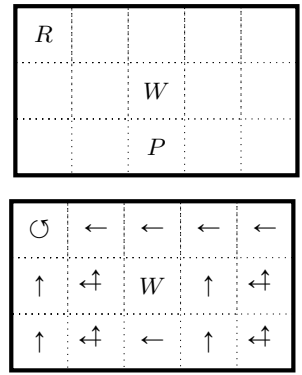

We benchmark our various FERL methods on the $3 \times 5$ grid-world problem \cite{sutton1990integrated} with an agent capable of taking the actions, up, down, left, right, or stand still, on a grid-world with one deterministic reward, one wall, and one penalty, as shown in Fig.~\ref{fig:grid-det} (top). The task is to find an optimal policy, as shown in Fig.~\ref{fig:grid-det} (bottom), for the agent at each state in the grid-world. 

\begin{figure}[b]
     \includegraphics[scale=.75]{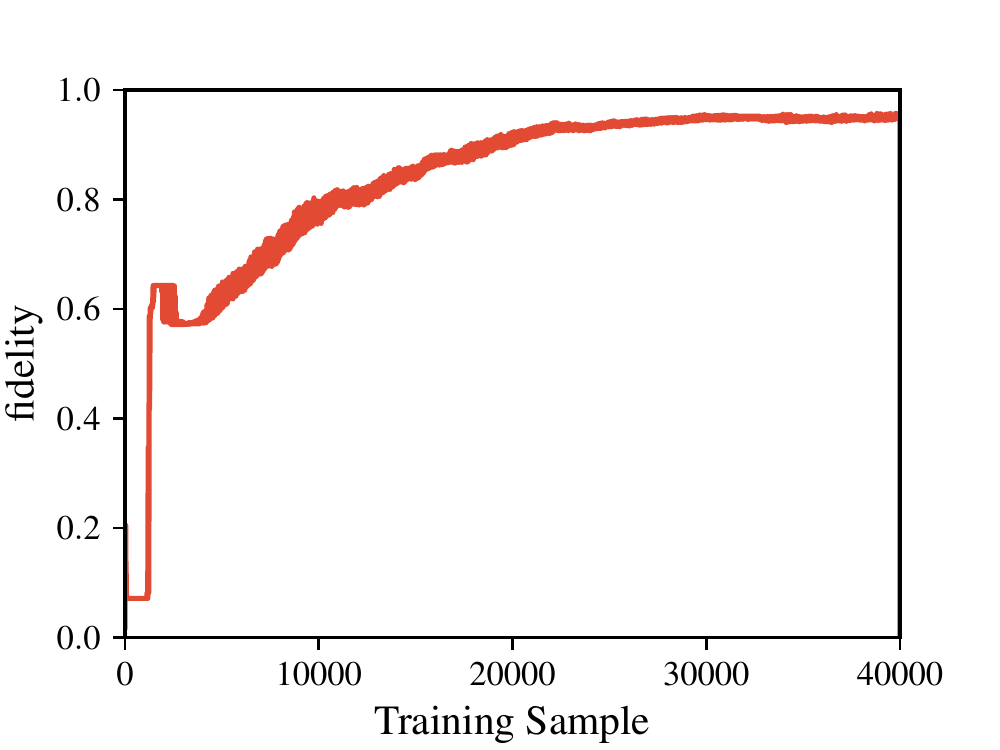}
\caption{\footnotesize{The learning curve of a deep \emph{Q}-network (DQN) with two hidden layers, each with eight hidden nodes, for the grid-world problem instance as shown in Fig.~\ref{fig:grid-det}.}}
  \label{fig:dqn}
\end{figure}

The discount factor, as explained in \S\ref{sec:MDP}, is set to $0.8$. The reward $R= 200$ is attained by the agent in the top-left corner, the neutral value of moving to any empty cell is $100$, and the agent is penalized by not receiving any reward if it moves to the penalty cell, with value $P= 0$. 

For $T_r$ independent runs of every FERL method, $T_s$ training samples are used. The fidelity measure at the $i$-th training sample is defined by
\begin{equation}
\text{fidelity}(i) =(T_r \times |S|)^{-1} \sum_{l = 1}^{T_r} \sum_{s\in S} \mathbbm{1}_{A(s, i, l) \in \pi^*(s)},  
\label{similaritymeasure}
\end{equation}
where $\pi^*$ denotes the best known policy and $A(s, i, l)$ denotes the action assigned at the $l$-th run and $i$-th training sample to the state $s$. In our experiments, each algorithm is run 100 times. 

Fig.~\ref{fig:dqn} demonstrates the performance of a deep \emph{Q}-network (DQN) network \cite{DQN} consisting of an input layer of 14 state nodes, two layers of eight hidden nodes each, and an output layer of five nodes representing the values of the \emph{Q}-function for different actions, given a configuration of state nodes. We use the same number of hidden nodes in the DQN as in the other networks described in this paper.

\subsection{Grid search for virtual parameters on the D-Wave 2000Q} \label{sec:gridsearch}

We treat the network on superconducting qubits represented in Fig.~\ref{fig:chimera} as a clamped QBM with two hidden layers, represented using blue and red colours. The state nodes are considered fully connected to the blue qubits and action nodes are fully connected to the red qubits. 

For a choice of virtual parameters $\Gamma \neq 0$ and $\beta$, which appear in \eqref{eff-hamiltonian} and \eqref{freeen-eff}, and for each query to the D-Wave 2000Q chip, we construct 150 effective classical configurations of one dimension higher, out of a pool of 3750 reads, according to the replica stacking method introduced in \S\ref{sec:stacking}. The 150 configurations are, in turn, employed to approximate the free energy of the quantum Hamiltonian. 
We conduct ten independent runs of FERL in this fashion and find the average fidelity over the ten runs and over the $T_s= 300$ training samples. 
\begin{figure}
  \centering
     \includegraphics[scale=0.75, trim=4mm 0 0 2mm]{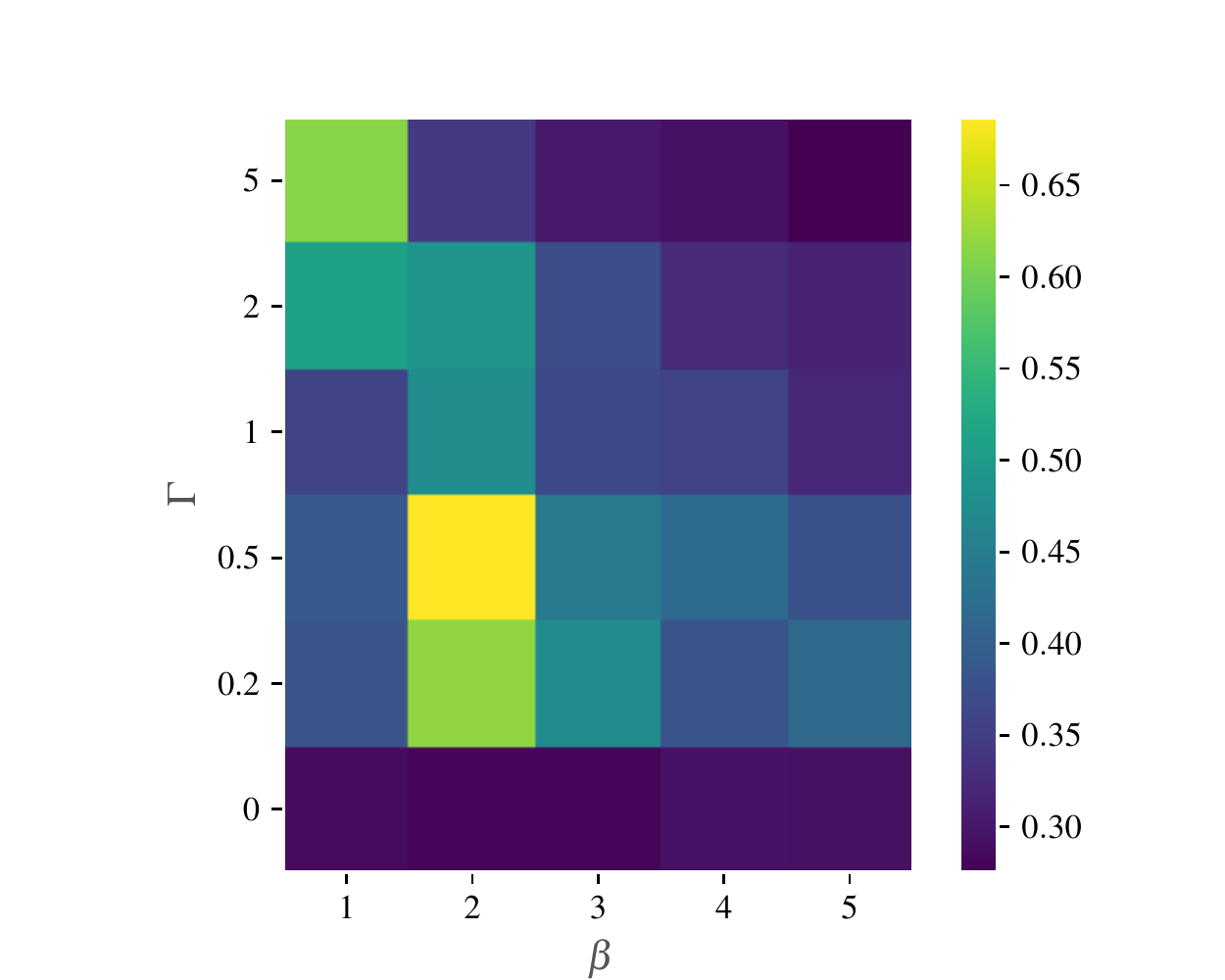}  
     \caption{\footnotesize{Heatmap of average fidelity obtained by various choices of virtual parameters $\beta$ and $\Gamma$. The $\Gamma = 0$ row tests the performance of FERL with samples obtained from the quantum annealer treated as classical configurations of a GBM. In all other rows, samples are interpreted as $\sigma^z$-measurements of a QBM.}}
  \label{fig:heatmap}
\end{figure}

Fig.~\ref{fig:heatmap} demonstrates a heatmap of average fidelity of each choice of virtual parameters $\beta$ and $\Gamma$. In the $\Gamma = 0$ row each D-Wave query is considered as sampling from a classical GBM with Fig.~\ref{fig:chimera} as underlying graph. 

\subsection{FERL for the grid-world problem} 

\begin{figure}
  \centering
         \includegraphics[scale=0.74, trim=7mm 0 0 -4mm]{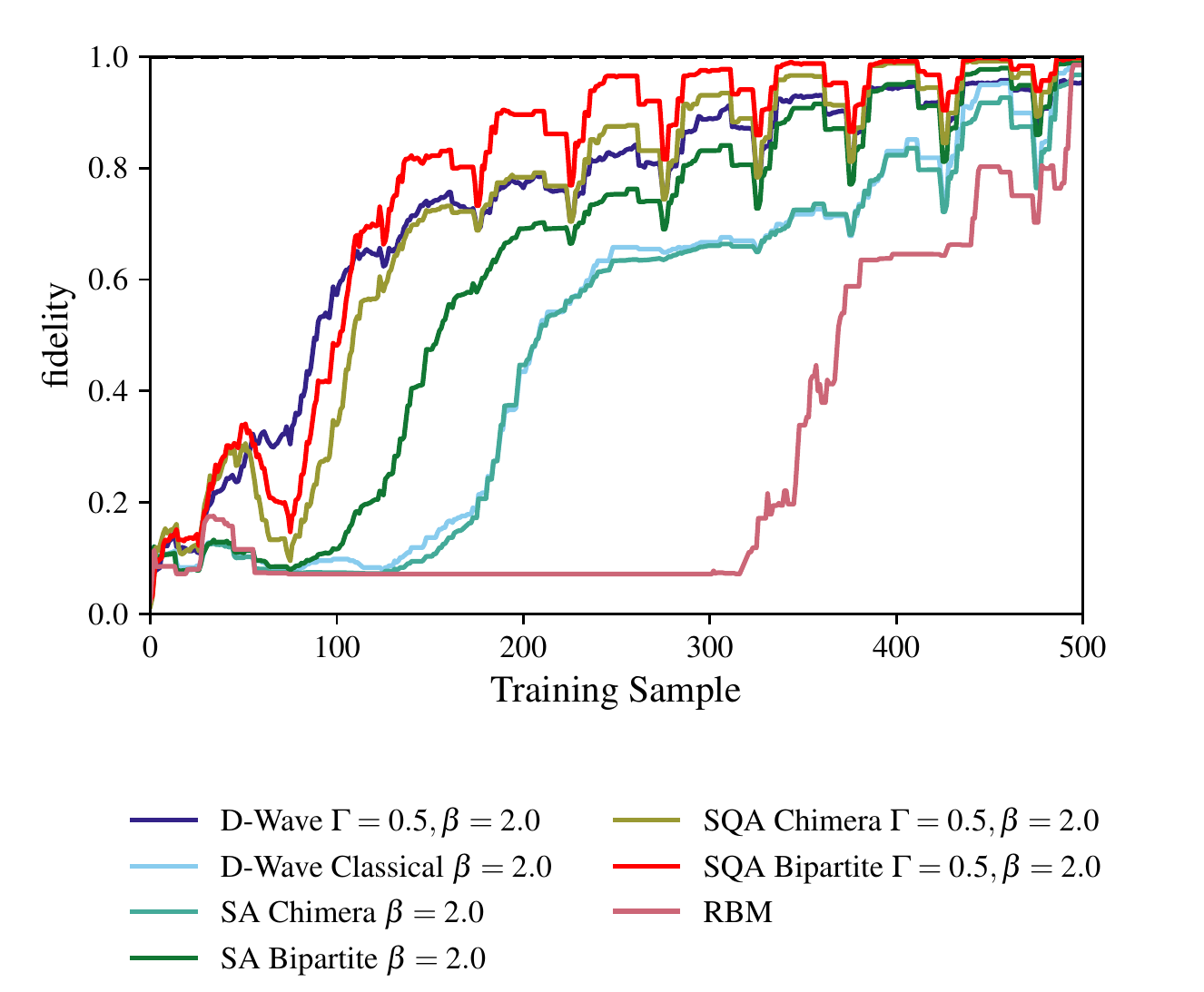}
  \caption{\footnotesize{Comparison of different FERL methods for the grid-world problem instance in Fig.~\ref{fig:grid-det}.}}
  \label{fig:ferl-comparison}
\end{figure}

Fig.~\ref{fig:ferl-comparison} shows the growth of the average fidelity of the best known policies generated by different FERL methods. For each method, the fidelity curve is an average over 100 independent runs, each for $T_s= 500$ training samples. 

In this figure, the ``D-Wave $\Gamma = 0.5$, $\beta = 2.0$'' curve corresponds to the D-Wave 2000Q replica stacking based method with the choice of best virtual parameters $\Gamma= 0.5$ and $\beta= 2.0$, as shown in the heatmap in Fig. \ref{fig:heatmap}. The training is based on formulae \eqref{q-vh-update}, \eqref{q-hh-update}, and \eqref{freeen-eff}. 
The ``SQA Bipartite $\Gamma = 0.5$, $\beta = 2.0$''  and ``SQA Chimera $\Gamma = 0.5$, $\beta = 2.0$'' curves are based on the same formulae with underlying graphs being a bipartite (DBM) and a Chimera graph, respectively, with the same choice of virtual parameters, but the effective Hamiltonian configurations generated using SQA as explained in \S\ref{sec:sqa}. 

The ``SA Bipartite $\beta = 2.0$'' and ``SA Chimera $\beta = 2.0$'' curves are generated by using SA to train a classical DBM and a classical GBM on the Chimera graph, respectively, using formulae \eqref{vh-update}, \eqref{hh-update}, and \eqref{freeen}. SA is run with a linear inverse temperature schedule, where $\beta = 2.0$ indicates the final value. The ``D-Wave Classical $\beta = 2.0$'' curve is generated using the same method, but with samples obtained using the D-Wave 2000Q. The ``RBM'' curve is generated using the method by \cite{hintonRBM}.

\section{Discussion} 

We solve the grid-world problem using various \emph{Q}-learning methods with the \emph{Q}-function parametrized by neural networks. For comparison, we demonstrate the performance of a DQN method that can be considered  state of the art. This method efficiently processes every training sample, but as shown in Fig.~\ref{fig:dqn}, requires very large number of training samples to converge to the optimal policy.  
Another conventional method is free energy-based reinforcement learning using an RBM. This method is also very successful at the scale of the reinforcement learning task considered in our experiment. Although this method is not outperforming other FERL methods taking advantage of a highly efficient sampling oracle,  the processing of each training sample is efficient as it is based on closed formulae. In fact, for the size of problem considered, the RBM-based FERL outperforms the DQN method.   

The comparison of results in Fig.~\ref{fig:ferl-comparison} suggests that replica stacking is a successful method for estimating effective classical configurations obtained from a quantum annealer, given that the spins can only be measured in measurement bases. For practical use in reinforcement learning, this method provides a means of treating the quantum annealer as a QBM. FERL using the quantum annealer, in conjunction with the replica stacking technique, provides significant improvement over FERL using classical Boltzmann machines. The curve representing SQA-based FERL using a Boltzmann machine on Chimera graph is almost coincident with the one obtained using the D-Wave 2000Q, whereas the SQA-based FERL using a DBM slightly outperforms it. This suggests that developing quantum annealing chips with greater connectivity and more control over annealing time can further improve the performance of the replica stacking method applied to reinforcement learning tasks. This is further supported by comparing the performance of SA-based FERL using a DBM versus SA-based FERL using the Chimera graph. This result shows that DBM is a better choice of neural network compared to the Chimera graph due to its additional connections.  

For practical reasons, we aim to associate an identical choice of virtual parameters $\beta$ and $\Gamma$ to all of the TFIMs constructed using FERL. \citet{PhysRevA.94.022308} and \citet{10.3389} provide methods for estimating the effective inverse temperature $\beta$ for other applications. However, in both studies, the samples obtained from the quantum annealer are matched to the Boltzmann distribution of a classical Ising model. In fact, the transverse-field strength is a second virtual parameter that we consider. The optimal choice $\Gamma = 0.5$ corresponds to $2/3$ of the annealing time, in agreement with the work of \cite{amin2015searching}, who also consider TFIM with 16 qubits. 

The agreement of FERL using quantum annealer reads treated as classical Boltzmann samples, with that of FERL using SA and classical Boltzmann machines, suggests that, at least for this task, and this size of Boltzmann machine, the   
measurements provided by the D-Wave 2000Q can be considered good approximations of Boltzmann distribution samples of classical Ising models.

\section{Conclusion} 

In this paper, we perform free energy-based reinforcement learning using existing quantum hardware, namely the D-Wave 2000Q. Our methods rely on Suzuki--Trotter decomposition and realization of the measured configurations by the quantum device as replicas of an effective classical Ising model of one dimension higher. Future research can employ these principles to solve larger-scale reinforcement learning tasks in the emerging field of quantum machine learning. 

{\bf Acknowledgement.} The authors would like to thank Marko~Bucyk for editing this manuscript.  

\small
\bibliography{ndp.bib}

\begin{thebibliography}{39}
\expandafter\ifx\csname natexlab\endcsname\relax\def\natexlab#1{#1}\fi
\expandafter\ifx\csname bibnamefont\endcsname\relax
  \def\bibnamefont#1{#1}\fi
\expandafter\ifx\csname bibfnamefont\endcsname\relax
  \def\bibfnamefont#1{#1}\fi
\expandafter\ifx\csname citenamefont\endcsname\relax
  \def\citenamefont#1{#1}\fi
\expandafter\ifx\csname url\endcsname\relax
  \def\url#1{\texttt{#1}}\fi
\expandafter\ifx\csname urlprefix\endcsname\relax\def\urlprefix{URL }\fi
\providecommand{\bibinfo}[2]{#2}
\providecommand{\eprint}[2][]{\url{#2}}

\bibitem[{\citenamefont{Sutton and Barto}(1998)}]{sutton-book}
\bibinfo{author}{\bibfnamefont{R.~S.} \bibnamefont{Sutton}} \bibnamefont{and}
  \bibinfo{author}{\bibfnamefont{A.~G.} \bibnamefont{Barto}},
  \emph{\bibinfo{title}{Reinforcement Learning : An Introduction}}
  (\bibinfo{publisher}{MIT Press}, \bibinfo{year}{1998}).

\bibitem[{\citenamefont{Bertsekas and Tsitsiklis}(1996)}]{bertsekas1996neuro}
\bibinfo{author}{\bibfnamefont{D.}~\bibnamefont{Bertsekas}} \bibnamefont{and}
  \bibinfo{author}{\bibfnamefont{J.}~\bibnamefont{Tsitsiklis}},
  \emph{\bibinfo{title}{Neuro-dynamic Programming}}, Anthropological Field
  Studies (\bibinfo{publisher}{Athena Scientific}, \bibinfo{year}{1996}), ISBN
  \bibinfo{isbn}{9781886529106},
  \urlprefix\url{https://books.google.ca/books?id=WxCCQgAACAAJ}.

\bibitem[{\citenamefont{Derhami et~al.}(2013)\citenamefont{Derhami,
  Khodadadian, Ghasemzadeh, and Bidoki}}]{derhami2013applying}
\bibinfo{author}{\bibfnamefont{V.}~\bibnamefont{Derhami}},
  \bibinfo{author}{\bibfnamefont{E.}~\bibnamefont{Khodadadian}},
  \bibinfo{author}{\bibfnamefont{M.}~\bibnamefont{Ghasemzadeh}},
  \bibnamefont{and} \bibinfo{author}{\bibfnamefont{A.~M.~Z.}
  \bibnamefont{Bidoki}}, \bibinfo{journal}{Applied Soft Computing}
  \textbf{\bibinfo{volume}{13}}, \bibinfo{pages}{1686} (\bibinfo{year}{2013}).

\bibitem[{\citenamefont{Syafiie et~al.}(2007)\citenamefont{Syafiie, Tadeo, and
  Martinez}}]{syafiie2007model}
\bibinfo{author}{\bibfnamefont{S.}~\bibnamefont{Syafiie}},
  \bibinfo{author}{\bibfnamefont{F.}~\bibnamefont{Tadeo}}, \bibnamefont{and}
  \bibinfo{author}{\bibfnamefont{E.}~\bibnamefont{Martinez}},
  \bibinfo{journal}{Engineering Applications of Artificial Intelligence}
  \textbf{\bibinfo{volume}{20}}, \bibinfo{pages}{767} (\bibinfo{year}{2007}).

\bibitem[{\citenamefont{Erev and Roth}(1998)}]{erev1998predicting}
\bibinfo{author}{\bibfnamefont{I.}~\bibnamefont{Erev}} \bibnamefont{and}
  \bibinfo{author}{\bibfnamefont{A.~E.} \bibnamefont{Roth}},
  \bibinfo{journal}{American Economic Review} pp. \bibinfo{pages}{848--881}
  (\bibinfo{year}{1998}).

\bibitem[{\citenamefont{Shteingart and
  Loewenstein}(2014)}]{shteingart2014reinforcement}
\bibinfo{author}{\bibfnamefont{H.}~\bibnamefont{Shteingart}} \bibnamefont{and}
  \bibinfo{author}{\bibfnamefont{Y.}~\bibnamefont{Loewenstein}},
  \bibinfo{journal}{Current Opinion in Neurobiology}
  \textbf{\bibinfo{volume}{25}}, \bibinfo{pages}{93} (\bibinfo{year}{2014}).

\bibitem[{\citenamefont{Matsui et~al.}(2011)\citenamefont{Matsui, Goto, Izumi,
  and Chen}}]{matsui2011compound}
\bibinfo{author}{\bibfnamefont{T.}~\bibnamefont{Matsui}},
  \bibinfo{author}{\bibfnamefont{T.}~\bibnamefont{Goto}},
  \bibinfo{author}{\bibfnamefont{K.}~\bibnamefont{Izumi}}, \bibnamefont{and}
  \bibinfo{author}{\bibfnamefont{Y.}~\bibnamefont{Chen}}, in
  \emph{\bibinfo{booktitle}{European Workshop on Reinforcement Learning}}
  (\bibinfo{organization}{Springer}, \bibinfo{year}{2011}), pp.
  \bibinfo{pages}{321--332}.

\bibitem[{\citenamefont{Sui et~al.}(2010)\citenamefont{Sui, Gosavi, and
  Lin}}]{sui2010reinforcement}
\bibinfo{author}{\bibfnamefont{Z.}~\bibnamefont{Sui}},
  \bibinfo{author}{\bibfnamefont{A.}~\bibnamefont{Gosavi}}, \bibnamefont{and}
  \bibinfo{author}{\bibfnamefont{L.}~\bibnamefont{Lin}},
  \bibinfo{journal}{Engineering Management Journal}
  \textbf{\bibinfo{volume}{22}}, \bibinfo{pages}{44} (\bibinfo{year}{2010}).

\bibitem[{\citenamefont{Sallans and Hinton}(2004)}]{hintonRBM}
\bibinfo{author}{\bibfnamefont{B.}~\bibnamefont{Sallans}} \bibnamefont{and}
  \bibinfo{author}{\bibfnamefont{G.~E.} \bibnamefont{Hinton}},
  \bibinfo{journal}{JMLR} \textbf{\bibinfo{volume}{5}}, \bibinfo{pages}{1063}
  (\bibinfo{year}{2004}).

\bibitem[{\citenamefont{Martens et~al.}(2013)\citenamefont{Martens,
  Chattopadhya, Pitassi, and Zemel}}]{martens2013representational}
\bibinfo{author}{\bibfnamefont{J.}~\bibnamefont{Martens}},
  \bibinfo{author}{\bibfnamefont{A.}~\bibnamefont{Chattopadhya}},
  \bibinfo{author}{\bibfnamefont{T.}~\bibnamefont{Pitassi}}, \bibnamefont{and}
  \bibinfo{author}{\bibfnamefont{R.}~\bibnamefont{Zemel}}, in
  \emph{\bibinfo{booktitle}{Advances in Neural Information Processing Systems}}
  (\bibinfo{year}{2013}), pp. \bibinfo{pages}{2877--2885}.

\bibitem[{\citenamefont{Hornik et~al.}(1989)\citenamefont{Hornik, Stinchcombe,
  and White}}]{hornik1989multilayer}
\bibinfo{author}{\bibfnamefont{K.}~\bibnamefont{Hornik}},
  \bibinfo{author}{\bibfnamefont{M.}~\bibnamefont{Stinchcombe}},
  \bibnamefont{and} \bibinfo{author}{\bibfnamefont{H.}~\bibnamefont{White}},
  \bibinfo{journal}{Neural Networks} \textbf{\bibinfo{volume}{2}},
  \bibinfo{pages}{359} (\bibinfo{year}{1989}).

\bibitem[{\citenamefont{Le~Roux and Bengio}(2008)}]{le2008representational}
\bibinfo{author}{\bibfnamefont{N.}~\bibnamefont{Le~Roux}} \bibnamefont{and}
  \bibinfo{author}{\bibfnamefont{Y.}~\bibnamefont{Bengio}},
  \bibinfo{journal}{Neural Computation} \textbf{\bibinfo{volume}{20}},
  \bibinfo{pages}{1631} (\bibinfo{year}{2008}).

\bibitem[{\citenamefont{{Crawford} et~al.}(2016)\citenamefont{{Crawford},
  {Levit}, {Ghadermarzy}, {Oberoi}, and {Ronagh}}}]{2016arXiv161205695C}
\bibinfo{author}{\bibfnamefont{D.}~\bibnamefont{{Crawford}}},
  \bibinfo{author}{\bibfnamefont{A.}~\bibnamefont{{Levit}}},
  \bibinfo{author}{\bibfnamefont{N.}~\bibnamefont{{Ghadermarzy}}},
  \bibinfo{author}{\bibfnamefont{J.~S.} \bibnamefont{{Oberoi}}},
  \bibnamefont{and} \bibinfo{author}{\bibfnamefont{P.}~\bibnamefont{{Ronagh}}},
  \bibinfo{journal}{ArXiv e-prints}  (\bibinfo{year}{2016}),
  \eprint{1612.05695}.

\bibitem[{\citenamefont{Yuksel}(2016)}]{Yuksel}
\bibinfo{author}{\bibfnamefont{S.}~\bibnamefont{Yuksel}}
  (\bibinfo{year}{2016}), \bibinfo{note}{course lecture notes, Queen's
  University (Kingston, ON Canada), Retrieved in May 2016},
  \urlprefix\url{http://www.mast.queensu.ca/~math472/Math472872LectureNotes.pdf}.

\bibitem[{\citenamefont{Bellman}(1956)}]{bellman1956dynamic}
\bibinfo{author}{\bibfnamefont{R.}~\bibnamefont{Bellman}},
  \bibinfo{journal}{Proceedings of the National Academy of Sciences}
  \textbf{\bibinfo{volume}{42}}, \bibinfo{pages}{767} (\bibinfo{year}{1956}).

\bibitem[{\citenamefont{Amin et~al.}(2016)\citenamefont{Amin, Andriyash, Rolfe,
  Kulchytskyy, and Melko}}]{1601.02036}
\bibinfo{author}{\bibfnamefont{M.~H.} \bibnamefont{Amin}},
  \bibinfo{author}{\bibfnamefont{E.}~\bibnamefont{Andriyash}},
  \bibinfo{author}{\bibfnamefont{J.}~\bibnamefont{Rolfe}},
  \bibinfo{author}{\bibfnamefont{B.}~\bibnamefont{Kulchytskyy}},
  \bibnamefont{and} \bibinfo{author}{\bibfnamefont{R.}~\bibnamefont{Melko}},
  \bibinfo{journal}{arXiv:1601.02036}  (\bibinfo{year}{2016}).

\bibitem[{\citenamefont{{Born} and {Fock}}(1928)}]{1928ZPhy}
\bibinfo{author}{\bibfnamefont{M.}~\bibnamefont{{Born}}} \bibnamefont{and}
  \bibinfo{author}{\bibfnamefont{V.}~\bibnamefont{{Fock}}},
  \bibinfo{journal}{Zeitschrift fur Physik} \textbf{\bibinfo{volume}{51}},
  \bibinfo{pages}{165} (\bibinfo{year}{1928}).

\bibitem[{\citenamefont{{Farhi} et~al.}(2000)\citenamefont{{Farhi},
  {Goldstone}, {Gutmann}, and {Sipser}}}]{2000quant.ph1106F}
\bibinfo{author}{\bibfnamefont{E.}~\bibnamefont{{Farhi}}},
  \bibinfo{author}{\bibfnamefont{J.}~\bibnamefont{{Goldstone}}},
  \bibinfo{author}{\bibfnamefont{S.}~\bibnamefont{{Gutmann}}},
  \bibnamefont{and} \bibinfo{author}{\bibfnamefont{M.}~\bibnamefont{{Sipser}}},
  \bibinfo{journal}{eprint arXiv:quant-ph/0001106}  (\bibinfo{year}{2000}),
  \eprint{quant-ph/0001106}.

\bibitem[{\citenamefont{Kadowaki and Nishimori}(1998)}]{PhysRevE.58.5355}
\bibinfo{author}{\bibfnamefont{T.}~\bibnamefont{Kadowaki}} \bibnamefont{and}
  \bibinfo{author}{\bibfnamefont{H.}~\bibnamefont{Nishimori}},
  \bibinfo{journal}{Phys. Rev. E} \textbf{\bibinfo{volume}{58}},
  \bibinfo{pages}{5355} (\bibinfo{year}{1998}),
  \urlprefix\url{https://link.aps.org/doi/10.1103/PhysRevE.58.5355}.

\bibitem[{\citenamefont{Johnson et~al.}(2011)\citenamefont{Johnson, Amin,
  Gildert, Lanting, Hamze, Dickson, Harris, Berkley, Johansson, Bunyk
  et~al.}}]{dwnature}
\bibinfo{author}{\bibfnamefont{M.~W.} \bibnamefont{Johnson}},
  \bibinfo{author}{\bibfnamefont{M.~H.~S.} \bibnamefont{Amin}},
  \bibinfo{author}{\bibfnamefont{S.}~\bibnamefont{Gildert}},
  \bibinfo{author}{\bibfnamefont{T.}~\bibnamefont{Lanting}},
  \bibinfo{author}{\bibfnamefont{F.}~\bibnamefont{Hamze}},
  \bibinfo{author}{\bibfnamefont{N.}~\bibnamefont{Dickson}},
  \bibinfo{author}{\bibfnamefont{R.}~\bibnamefont{Harris}},
  \bibinfo{author}{\bibfnamefont{A.~J.} \bibnamefont{Berkley}},
  \bibinfo{author}{\bibfnamefont{J.}~\bibnamefont{Johansson}},
  \bibinfo{author}{\bibfnamefont{P.}~\bibnamefont{Bunyk}},
  \bibnamefont{et~al.}, \bibinfo{journal}{Nature}
  \textbf{\bibinfo{volume}{473}}, \bibinfo{pages}{194} (\bibinfo{year}{2011}),
  \urlprefix\url{http://dx.doi.org/10.1038/nature10012}.

\bibitem[{\citenamefont{Sarandy and Lidar}(2005)}]{PhysRevA.71.012331}
\bibinfo{author}{\bibfnamefont{M.~S.} \bibnamefont{Sarandy}} \bibnamefont{and}
  \bibinfo{author}{\bibfnamefont{D.~A.} \bibnamefont{Lidar}},
  \bibinfo{journal}{Phys. Rev. A} \textbf{\bibinfo{volume}{71}},
  \bibinfo{pages}{012331} (\bibinfo{year}{2005}),
  \urlprefix\url{https://link.aps.org/doi/10.1103/PhysRevA.71.012331}.

\bibitem[{\citenamefont{Venuti et~al.}(2016)\citenamefont{Venuti, Albash,
  Lidar, and Zanardi}}]{PhysRevA.93.032118}
\bibinfo{author}{\bibfnamefont{L.~C.} \bibnamefont{Venuti}},
  \bibinfo{author}{\bibfnamefont{T.}~\bibnamefont{Albash}},
  \bibinfo{author}{\bibfnamefont{D.~A.} \bibnamefont{Lidar}}, \bibnamefont{and}
  \bibinfo{author}{\bibfnamefont{P.}~\bibnamefont{Zanardi}},
  \bibinfo{journal}{Phys. Rev. A} \textbf{\bibinfo{volume}{93}},
  \bibinfo{pages}{032118} (\bibinfo{year}{2016}),
  \urlprefix\url{http://link.aps.org/doi/10.1103/PhysRevA.93.032118}.

\bibitem[{\citenamefont{Albash et~al.}(2012)\citenamefont{Albash, Boixo, Lidar,
  and Zanardi}}]{136726301412123016}
\bibinfo{author}{\bibfnamefont{T.}~\bibnamefont{Albash}},
  \bibinfo{author}{\bibfnamefont{S.}~\bibnamefont{Boixo}},
  \bibinfo{author}{\bibfnamefont{D.~A.} \bibnamefont{Lidar}}, \bibnamefont{and}
  \bibinfo{author}{\bibfnamefont{P.}~\bibnamefont{Zanardi}},
  \bibinfo{journal}{New Journal of Physics} \textbf{\bibinfo{volume}{14}},
  \bibinfo{pages}{123016} (\bibinfo{year}{2012}),
  \urlprefix\url{http://stacks.iop.org/1367-2630/14/i=12/a=123016}.

\bibitem[{\citenamefont{Avron et~al.}(2012)\citenamefont{Avron, Fraas, Graf,
  and Grech}}]{Avron2012}
\bibinfo{author}{\bibfnamefont{J.~E.} \bibnamefont{Avron}},
  \bibinfo{author}{\bibfnamefont{M.}~\bibnamefont{Fraas}},
  \bibinfo{author}{\bibfnamefont{G.~M.} \bibnamefont{Graf}}, \bibnamefont{and}
  \bibinfo{author}{\bibfnamefont{P.}~\bibnamefont{Grech}},
  \bibinfo{journal}{Communications in Mathematical Physics}
  \textbf{\bibinfo{volume}{314}}, \bibinfo{pages}{163} (\bibinfo{year}{2012}),
  ISSN \bibinfo{issn}{1432-0916},
  \urlprefix\url{http://dx.doi.org/10.1007/s00220-012-1504-1}.

\bibitem[{\citenamefont{{Bachmann} et~al.}(2016)\citenamefont{{Bachmann}, {De
  Roeck}, and {Fraas}}}]{2016arXiv161201505B}
\bibinfo{author}{\bibfnamefont{S.}~\bibnamefont{{Bachmann}}},
  \bibinfo{author}{\bibfnamefont{W.}~\bibnamefont{{De Roeck}}},
  \bibnamefont{and} \bibinfo{author}{\bibfnamefont{M.}~\bibnamefont{{Fraas}}},
  \bibinfo{journal}{ArXiv e-prints}  (\bibinfo{year}{2016}),
  \eprint{1612.01505}.

\bibitem[{\citenamefont{Amin}(2015)}]{amin2015searching}
\bibinfo{author}{\bibfnamefont{M.~H.} \bibnamefont{Amin}},
  \bibinfo{journal}{Phys. Rev. A} \textbf{\bibinfo{volume}{92}},
  \bibinfo{pages}{052323} (\bibinfo{year}{2015}).

\bibitem[{\citenamefont{Suzuki}(1976)}]{suzuki1976relationship}
\bibinfo{author}{\bibfnamefont{M.}~\bibnamefont{Suzuki}},
  \bibinfo{journal}{Progress of Theoretical Physics}
  \textbf{\bibinfo{volume}{56}}, \bibinfo{pages}{1454} (\bibinfo{year}{1976}).

\bibitem[{\citenamefont{Anthony~Brabazon}(2015)}]{naturalcomputingalgorithms}
\bibinfo{author}{\bibfnamefont{S.~M.} \bibnamefont{Anthony~Brabazon},
  \bibfnamefont{Michael~O'Neill}}, \emph{\bibinfo{title}{Natural Computing
  Algorithms}} (\bibinfo{publisher}{Springer-Verlag Berlin Heidelberg},
  \bibinfo{year}{2015}).

\bibitem[{\citenamefont{Marto{\v{n}}{\'a}k
  et~al.}(2002)\citenamefont{Marto{\v{n}}{\'a}k, Santoro, and
  Tosatti}}]{martovnak2002quantum}
\bibinfo{author}{\bibfnamefont{R.}~\bibnamefont{Marto{\v{n}}{\'a}k}},
  \bibinfo{author}{\bibfnamefont{G.~E.} \bibnamefont{Santoro}},
  \bibnamefont{and} \bibinfo{author}{\bibfnamefont{E.}~\bibnamefont{Tosatti}},
  \bibinfo{journal}{Phys. Rev. B} \textbf{\bibinfo{volume}{66}},
  \bibinfo{pages}{094203} (\bibinfo{year}{2002}).

\bibitem[{\citenamefont{Heim et~al.}(2015)\citenamefont{Heim, R{\o}nnow,
  Isakov, and Troyer}}]{1411.5693}
\bibinfo{author}{\bibfnamefont{B.}~\bibnamefont{Heim}},
  \bibinfo{author}{\bibfnamefont{T.~F.} \bibnamefont{R{\o}nnow}},
  \bibinfo{author}{\bibfnamefont{S.~V.} \bibnamefont{Isakov}},
  \bibnamefont{and} \bibinfo{author}{\bibfnamefont{M.}~\bibnamefont{Troyer}},
  \bibinfo{journal}{Science} \textbf{\bibinfo{volume}{348}},
  \bibinfo{pages}{215} (\bibinfo{year}{2015}).

\bibitem[{\citenamefont{Isakov et~al.}(2015)\citenamefont{Isakov, Mazzola,
  Smelyanskiy, Jiang, Boixo, Neven, and Troyer}}]{1510.08057}
\bibinfo{author}{\bibfnamefont{S.~V.} \bibnamefont{Isakov}},
  \bibinfo{author}{\bibfnamefont{G.}~\bibnamefont{Mazzola}},
  \bibinfo{author}{\bibfnamefont{V.~N.} \bibnamefont{Smelyanskiy}},
  \bibinfo{author}{\bibfnamefont{Z.}~\bibnamefont{Jiang}},
  \bibinfo{author}{\bibfnamefont{S.}~\bibnamefont{Boixo}},
  \bibinfo{author}{\bibfnamefont{H.}~\bibnamefont{Neven}}, \bibnamefont{and}
  \bibinfo{author}{\bibfnamefont{M.}~\bibnamefont{Troyer}},
  \bibinfo{journal}{arXiv:1510.08057}  (\bibinfo{year}{2015}).

\bibitem[{\citenamefont{{Albash} et~al.}(2014)\citenamefont{{Albash},
  {R{\o}nnow}, {Troyer}, and {Lidar}}}]{2014arXiv1409.3827A}
\bibinfo{author}{\bibfnamefont{T.}~\bibnamefont{{Albash}}},
  \bibinfo{author}{\bibfnamefont{T.~F.} \bibnamefont{{R{\o}nnow}}},
  \bibinfo{author}{\bibfnamefont{M.}~\bibnamefont{{Troyer}}}, \bibnamefont{and}
  \bibinfo{author}{\bibfnamefont{D.~A.} \bibnamefont{{Lidar}}},
  \bibinfo{journal}{ArXiv e-prints}  (\bibinfo{year}{2014}),
  \eprint{1409.3827}.

\bibitem[{\citenamefont{Brady and van Dam}(2016)}]{arXiv:1509.02562v2}
\bibinfo{author}{\bibfnamefont{L.~T.} \bibnamefont{Brady}} \bibnamefont{and}
  \bibinfo{author}{\bibfnamefont{W.}~\bibnamefont{van Dam}},
  \bibinfo{journal}{Phys. Rev. A} \textbf{\bibinfo{volume}{93}},
  \bibinfo{pages}{032304} (\bibinfo{year}{2016}).

\bibitem[{\citenamefont{{Shin} et~al.}(2014)\citenamefont{{Shin}, {Smith},
  {Smolin}, and {Vazirani}}}]{arXiv:1401.7087v2}
\bibinfo{author}{\bibfnamefont{S.~W.} \bibnamefont{{Shin}}},
  \bibinfo{author}{\bibfnamefont{G.}~\bibnamefont{{Smith}}},
  \bibinfo{author}{\bibfnamefont{J.~A.} \bibnamefont{{Smolin}}},
  \bibnamefont{and}
  \bibinfo{author}{\bibfnamefont{U.}~\bibnamefont{{Vazirani}}},
  \bibinfo{journal}{ArXiv e-prints}  (\bibinfo{year}{2014}),
  \eprint{1401.7087}.

\bibitem[{\citenamefont{Morita and Nishimori}(2006)}]{morita2006convergence}
\bibinfo{author}{\bibfnamefont{S.}~\bibnamefont{Morita}} \bibnamefont{and}
  \bibinfo{author}{\bibfnamefont{H.}~\bibnamefont{Nishimori}},
  \bibinfo{journal}{Journal of Physics A: Mathematical and General}
  \textbf{\bibinfo{volume}{39}}, \bibinfo{pages}{13903} (\bibinfo{year}{2006}).

\bibitem[{\citenamefont{Sutton}(1990)}]{sutton1990integrated}
\bibinfo{author}{\bibfnamefont{R.~S.} \bibnamefont{Sutton}}, in
  \emph{\bibinfo{booktitle}{In Proceedings of the Seventh International
  Conference on Machine Learning}} (\bibinfo{publisher}{Morgan Kaufmann},
  \bibinfo{year}{1990}), pp. \bibinfo{pages}{216--224}.

\bibitem[{\citenamefont{Mnih et~al.}(2015)\citenamefont{Mnih, Kavukcuoglu,
  Silver, Rusu, Veness, Bellemare, Graves, Riedmiller, Fidjeland, Ostrovski
  et~al.}}]{DQN}
\bibinfo{author}{\bibfnamefont{V.}~\bibnamefont{Mnih}},
  \bibinfo{author}{\bibfnamefont{K.}~\bibnamefont{Kavukcuoglu}},
  \bibinfo{author}{\bibfnamefont{D.}~\bibnamefont{Silver}},
  \bibinfo{author}{\bibfnamefont{A.~A.} \bibnamefont{Rusu}},
  \bibinfo{author}{\bibfnamefont{J.}~\bibnamefont{Veness}},
  \bibinfo{author}{\bibfnamefont{M.~G.} \bibnamefont{Bellemare}},
  \bibinfo{author}{\bibfnamefont{A.}~\bibnamefont{Graves}},
  \bibinfo{author}{\bibfnamefont{M.}~\bibnamefont{Riedmiller}},
  \bibinfo{author}{\bibfnamefont{A.~K.} \bibnamefont{Fidjeland}},
  \bibinfo{author}{\bibfnamefont{G.}~\bibnamefont{Ostrovski}},
  \bibnamefont{et~al.}, \bibinfo{journal}{Nature}
  \textbf{\bibinfo{volume}{518}}, \bibinfo{pages}{529} (\bibinfo{year}{2015}),
  \urlprefix\url{http://dx.doi.org/10.1038/nature14236}.

\bibitem[{\citenamefont{Benedetti et~al.}(2016)\citenamefont{Benedetti,
  Realpe-G\'omez, Biswas, and Perdomo-Ortiz}}]{PhysRevA.94.022308}
\bibinfo{author}{\bibfnamefont{M.}~\bibnamefont{Benedetti}},
  \bibinfo{author}{\bibfnamefont{J.}~\bibnamefont{Realpe-G\'omez}},
  \bibinfo{author}{\bibfnamefont{R.}~\bibnamefont{Biswas}}, \bibnamefont{and}
  \bibinfo{author}{\bibfnamefont{A.}~\bibnamefont{Perdomo-Ortiz}},
  \bibinfo{journal}{Phys. Rev. A} \textbf{\bibinfo{volume}{94}},
  \bibinfo{pages}{022308} (\bibinfo{year}{2016}),
  \urlprefix\url{https://link.aps.org/doi/10.1103/PhysRevA.94.022308}.

\bibitem[{\citenamefont{Raymond et~al.}(2016)\citenamefont{Raymond, Yarkoni,
  and Andriyash}}]{10.3389}
\bibinfo{author}{\bibfnamefont{J.}~\bibnamefont{Raymond}},
  \bibinfo{author}{\bibfnamefont{S.}~\bibnamefont{Yarkoni}}, \bibnamefont{and}
  \bibinfo{author}{\bibfnamefont{E.}~\bibnamefont{Andriyash}},
  \bibinfo{journal}{Frontiers in ICT} \textbf{\bibinfo{volume}{3}},
  \bibinfo{pages}{23} (\bibinfo{year}{2016}), ISSN \bibinfo{issn}{2297-198X},
  \urlprefix\url{http://journal.frontiersin.org/article/10.3389/fict.2016.00023}.

\end{thebibliography}

\end{document}